\documentclass[conference]{IEEEtran}
\IEEEoverridecommandlockouts
\usepackage{cite}
\usepackage{amsmath,amssymb,amsfonts}
\usepackage{algorithmic}
\usepackage{graphicx}
\usepackage{textcomp}
\usepackage{xcolor}
\usepackage{booktabs}

\def\BibTeX{{\rm B\kern-.05em{\sc i\kern-.025em b}\kern-.08em
    T\kern-.1667em\lower.7ex\hbox{E}\kern-.125emX}}
\begin{document}

\title{REALMS: An AI-Assistant Conversational System for Real-Time Exact Audience Sizing over High-Dimensional Nested Profiles\\
% {\footnotesize \textsuperscript{*}Note: Sub-titles are not captured for https://ieeexplore.ieee.org  and
}

% \author{
% % \IEEEauthorblockN{Haixu Ma}
% % \IEEEauthorblockA{\textit{dept. name of organization (of Aff.)} \\
% % \textit{name of organization (of Aff.)}\\
% % City, Country \\
% % email address or ORCID}

% \IEEEauthorblockN{Haixu Ma}
% % \IEEEauthorblockA{\textit{dept. name of organization (of Aff.)} \\
% \textit{Adobe Inc.}\\
% San Jose, USA \\
% haixuma@gmail.com

% \and
% % \IEEEauthorblockN{2\textsuperscript{nd} Given Name Surname}
% % \IEEEauthorblockA{\textit{dept. name of organization (of Aff.)} \\
% % \textit{name of organization (of Aff.)}\\
% % City, Country \\
% % email address or ORCID}

% \IEEEauthorblockN{Aditya Bansal}
% % \IEEEauthorblockA{\textit{dept. name of organization (of Aff.)} \\
% \textit{Adobe Inc.}\\
% San Jose, USA \\
% adibansal@adobe.com

% \and
% % \IEEEauthorblockN{3\textsuperscript{rd} Given Name Surname}
% % \IEEEauthorblockA{\textit{dept. name of organization (of Aff.)} \\
% % \textit{name of organization (of Aff.)}\\
% % City, Country \\
% % email address or ORCID}

% \IEEEauthorblockN{Shubham Lohiya}
% % \IEEEauthorblockA{\textit{dept. name of organization (of Aff.)} \\
% \textit{Adobe Inc.}\\
% San Jose, USA \\
% slohiya@adobe.com

% \and

% \IEEEauthorblockN{Sumit Ranjan}
% % \IEEEauthorblockA{\textit{dept. name of organization (of Aff.)} \\
% \textit{Adobe Inc.}\\
% San Jose, USA \\
% sumit.nitt@gmail.com

% }

\author{
\IEEEauthorblockN{Haixu Ma\textsuperscript{*}\thanks{*All authors contributed equally.}}
\IEEEauthorblockA{
\textit{Adobe Inc.}\\
San Jose, USA\\
haixuma@gmail.com
}

\and

\IEEEauthorblockN{Aditya Bansal\textsuperscript{*}}
\IEEEauthorblockA{
\textit{Adobe Inc.}\\
San Jose, USA\\
adibansal@adobe.com
}

\and

\IEEEauthorblockN{Shubham Lohiya\textsuperscript{*}}
\IEEEauthorblockA{
\textit{Adobe Inc.}\\
San Jose, USA\\
slohiya@adobe.com
}

\and

\IEEEauthorblockN{Sumit Ranjan\textsuperscript{*}}
\IEEEauthorblockA{
\textit{Adobe Inc.}\\
San Jose, USA\\
sumit.nitt@gmail.com
}
}

\maketitle

\begin{abstract}
 Audience sizing is a critical component of digital marketing. It enables precise resource allocation, campaign planning, and performance optimization. Traditional approaches using skeleton audiences, sampling, or predictive modeling suffer from significant delays, estimation errors, and poor scalability over high-dimensional profile data. We present REALMS (\textbf{R}eal-time \textbf{E}xact \textbf{A}udience sizing via \textbf{L}LM-based \textbf{M}ulti-attribute \textbf{S}earch), a conversational system for exact audience sizing deployed in production on an enterprise customer data platform. REALMS enables marketers to query massive profile stores with millions of profiles and thousands of attributes using natural language and receive precise counts in seconds. The system introduces three key components: (1) a categorical attribute retrieval mechanism using embedding-based vector search to dynamically identify relevant schema attributes without manual configuration; (2) an LLM-powered NL2SQL pipeline with template-based in-context learning for accurate query generation over complex nested schemas; and (3) schema standardization enabling industry-agnostic deployment across diverse enterprise environments. Evaluation on real enterprise data demonstrates strong recall for attribute retrieval, high SQL execution accuracy, and low latency, which enables real-time interactive audience insights where prior methods required hours.
\end{abstract}


\noindent
\section{Introduction}

Digital marketing platforms rely on audience sizing to estimate how many users satisfy behavioral, demographic, and transactional constraints prior to campaign deployment. Accurate estimates support campaign planning, budget allocation, and personalization, whereas inaccurate estimates can lead to inefficient spending and degraded outcomes. The increasing availability of large-scale user data has further strengthened the role of analytics in enterprise decision making \cite{wedel2016marketing, chen2012business}.

In enterprise customer data platforms, audience sizing is often implemented via approximate query processing, such as sampling, sketches, or predictive modeling. Although these techniques offer established efficiency--accuracy trade-offs \cite{chaudhuri1998random, cormode2011synopses}, they can be brittle in practice: estimation error is difficult to control for correlated, high-dimensional, and nested attributes, and batch-oriented pre-computation can inflate end-to-end latency, limiting iterative and interactive use.

In parallel, natural language interfaces have become an important access modality for non-technical data exploration. Conversational information seeking systems reduce the need to manually author structured queries \cite{radlinski2017theoretical, dalton2022conversational}. Audience sizing, however, differs from conventional conversational information retrieval because it requires executing aggregation queries over structured profile stores rather than retrieving ranked documents, necessitating robust intent understanding coupled with reliable structured execution. Recent advances in Large Language Models (LLMs) have improved Natural-Language-to-SQL (NL2SQL) and semantic parsing \cite{zhong2017seq2sql, yu2018spider, sharma2025ttd}, but deploying these methods for enterprise audience sizing raises additional requirements beyond standard benchmark settings.

We summarize the key requirements that shape our system design as follows: 

(1) \textbf{Adaptability to high-dimensional schemas and heterogeneous data sources}: Enterprise profile schemas are highly diverse and often contain thousands of attributes organized in nested hierarchies. In addition, customer data platforms integrate multiple sources, such as web interactions, transactions, and offline records, with differing representations and semantics. A practical solution should therefore perform robust schema grounding and schema alignment, mapping natural-language constraints to the appropriate structured attributes without manual rules or per-customer customization; 

(2) \textbf{Scalability over large profile datasets}: Audience sizing operates over very large profile stores, where high dimensionality and nested structures make exact counting computationally expensive. The system should employ efficient retrieval and execution strategies to support exact computation while controlling resource consumption; 

(3) \textbf{Real-time conversational interaction}: In an AI-assistant setting, end-to-end latency directly affects usability. The system should reliably produce executable queries and return precise results within seconds to preserve interactive dialogue.

\begin{figure}[t]
    \centering
    \includegraphics[width=1\linewidth]{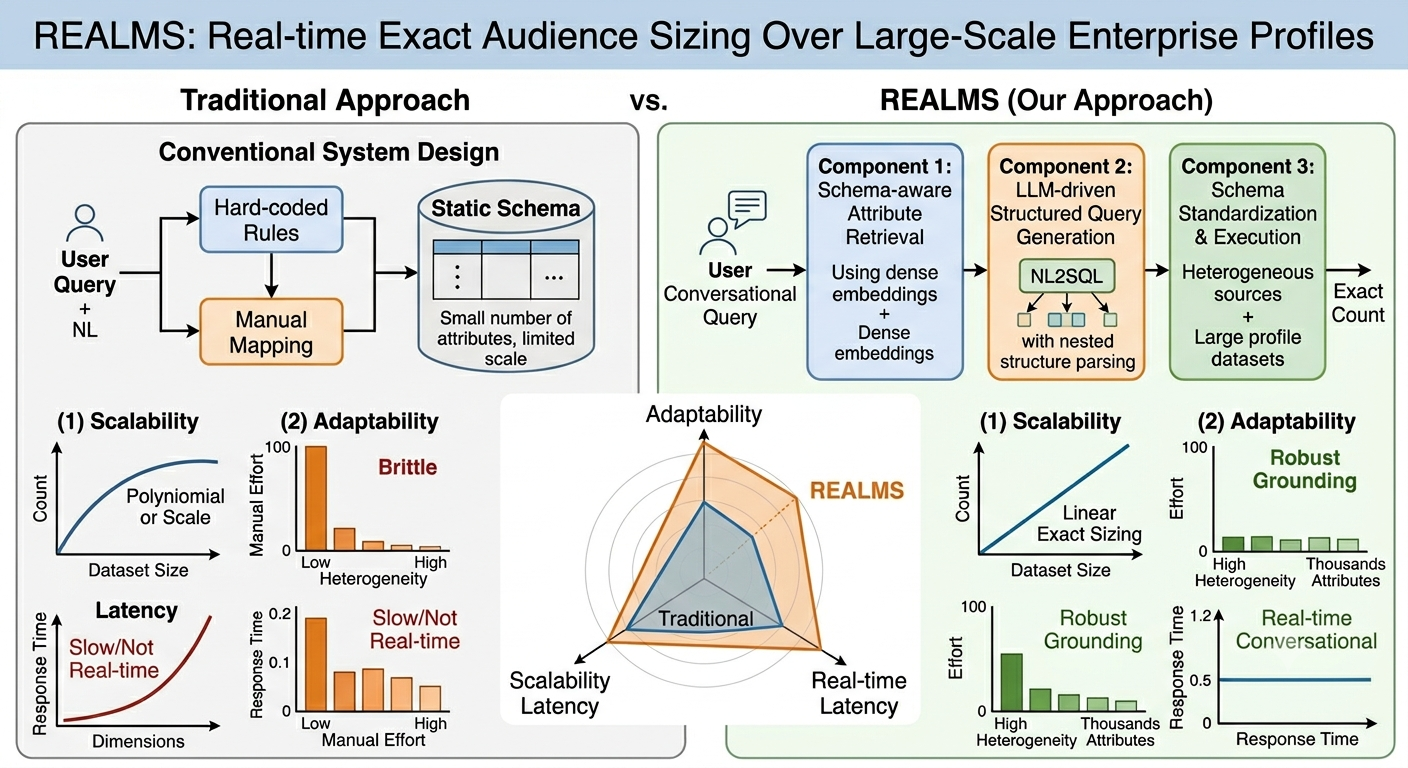}
    \caption{Comparative Analysis of Conventional and REALMS (Our Approach) for Real-time Exact Audience Sizing Over Large-Scale Enterprise Profiles.}
    \label{fig:architecture}
\end{figure}

To address these challenges, we present \textbf{REALMS} (Real-time Exact Audience sizing via LLM-based Multi-attribute Search), a conversational system for natural-language audience sizing over large-scale enterprise profile datasets. REALMS integrates information retrieval, semantic parsing, and database execution within an AI assistant. The system comprises three components: 

(1) \textbf{Schema-aware attribute retrieval}, which uses embedding-based dense retrieval to map user utterances to relevant schema attributes \cite{reimers2019sentence, karpukhin2020dense}; 

(2) \textbf{LLM-driven structured query generation}, which produces executable NL2SQL queries over deeply nested schemas via in-context generation; 

(3) \textbf{Schema standardization and execution}, which normalizes heterogeneous data sources and supports exact counting at scale.

\begin{table}[t]
\centering
\caption{Comparison of audience sizing approaches.}
\label{tab:comparison_methods}
\small
\setlength{\tabcolsep}{4pt}
\renewcommand{\arraystretch}{1.05}
\begin{tabular}{lcccc}
\toprule
 & Skeleton & Sampling & Predictive & REALMS \\
 & Audiences & Techniques & Models & \textbf{(Ours)} \\
\midrule
\textbf{Real-time}      & \texttimes & \texttimes & \texttimes & \checkmark \\
\textbf{Exact sizing}   & \checkmark & \texttimes & \texttimes & \checkmark \\
\textbf{Compute needed} & High       & Medium     & High       & Medium \\
\bottomrule
\end{tabular}
\end{table}

Our experiments on real enterprise datasets show high attribute-retrieval recall, strong SQL execution accuracy, and low end-to-end latency, enabling interactive audience insights in seconds where conventional pipelines may require minutes to hours. More broadly, REALMS exemplifies a conversational analytics paradigm in which an LLM-based assistant functions as a semantic query planner that bridges information retrieval and database execution: it retrieves relevant schema elements and composes structured aggregation queries to return exact audience counts. We further demonstrate practicality through a production deployment in an enterprise conversational AI assistant.

\section{Related Work}

\textbf{Conversational Information Retrieval.} Foundational work on conversational search~\cite{radlinski2017theoretical} and conversational information seeking~\cite{zamani2022conversational} investigates multi-turn dialogue for complex information needs. Liu et al.~\cite{liu2024suql} proposed SUQL, augmenting SQL with free-text primitives for conversational search over hybrid data. While prior conversational systems primarily retrieve ranked documents~\cite{dalton2022conversational}, our work retrieves structured schema elements and computes exact aggregation queries, extending conversational IR to structured enterprise analytics.

\textbf{Natural Language Interfaces to Databases (NL2SQL).} Semantic parsing into SQL has advanced through benchmarks such as WikiSQL~\cite{zhong2017seq2sql}, Spider~\cite{yu2018spider}, and BIRD~\cite{li2024bird}, which evaluates text-to-SQL over large-scale databases with noisy data. LLMs with in-context learning achieve competitive accuracy without fine-tuning~\cite{rajkumar2022evaluating}, with DIN-SQL~\cite{pourreza2023dinsql} and DAIL-SQL~\cite{gao2024dailsql} further advancing prompt-based generation through decomposed reasoning and optimized prompt design. However, existing work predominantly assumes curated schemas of moderate size, whereas enterprise profile stores contain thousands of heterogeneous, deeply nested attributes requiring real-time interactive querying.

\textbf{LLM-based Enterprise Analytics Systems.} SiriusBI~\cite{jiang2024siriusbi} deploys an LLM-based business intelligence system with multi-round dialogue and schema-aware prompting. Floratou et al.~\cite{floratou2024nl2sql} identify persistent production challenges including schema complexity and execution reliability. Sharma et al.~\cite{sharma2025ttd} propose tree-guided token decoding for schema-aware SQL generation. Our system addresses these challenges through template-based in-context learning, categorical attribute retrieval, and schema standardization.

\textbf{Schema Matching and Data Integration.} Classical schema matching~\cite{rahm2001survey} and data integration~\cite{doan2012principles} address aligning heterogeneous sources into unified queryable representations. Recent work applies LLMs: ReMatch~\cite{sheetrit2024rematch} employs retrieval-enhanced LLMs for large-scale schema matching, and CRUSH4SQL~\cite{kothyari2023crush4sql} uses LLM-generated schema hallucinations with composite dense retrieval to identify relevant schema subsets. Our system similarly leverages embedding-based retrieval but introduces categorical bucketing for diverse and precise attribute coverage.

\section{System Design}
To address the challenges of schema heterogeneity, large-scale profile data, and interactive latency outlined in Section~1, REALMS is designed around three core principles that jointly enable accurate and efficient audience sizing from natural-language requests.

First, REALMS adopts \emph{schema-agnostic attribute retrieval} to bridge the gap between user language and highly diverse enterprise schemas. Enterprise profile stores often contain thousands of attributes organized in deeply nested structures, with naming conventions and semantics that vary substantially across customers and data sources. To address this challenge, REALMS leverages dense embedding representations combined with categorical attribute bucketing to encode schema semantics into a shared embedding space~\cite{reimers2019sentence}. At query time, natural-language constraints are matched against candidate attributes through semantic retrieval rather than hand-crafted rules or schema-specific mappings. This design allows the system to generalize across heterogeneous schemas while minimizing customer-specific engineering effort.

Second, REALMS employs \emph{template-based in-context learning} for robust NL2SQL generation~\cite{pourreza2023dinsql, gao2024dailsql}. Instead of relying on task-specific fine-tuning for each customer schema, the system constructs prompts using retrieved schema context and reusable query templates. Large language models then translate natural-language audience definitions into executable SQL queries over complex nested profile structures. This approach improves portability across domains while maintaining the flexibility needed to support diverse audience construction tasks.

Third, REALMS introduces a \emph{schema standardization layer} that provides a unified execution interface over heterogeneous enterprise data sources. Customer data platforms typically integrate information from web interactions, transactions, and offline records, each with different schemas and storage formats. REALMS materializes these data into a standardized columnar representation that abstracts away source-specific differences and exposes a consistent query surface. By decoupling query generation from physical data organization, the standardization layer enables efficient execution and exact audience counting at scale while preserving compatibility with existing customer data infrastructures.

Together, these design principles establish the foundation of REALMS, enabling semantic understanding of natural-language requests, portable query generation across heterogeneous schemas, and scalable exact computation suitable for real-time conversational interaction. Below is a detailed introduction for the architexture of our system.

The high-level architecture is illustrated in Figure~\ref{fig:architecture}. The system comprises two stages: an \textit{offline pre-processing} stage that prepares and indexes enterprise profile data, and an \textit{AI Assistant runtime} that handles real-time query processing. This separation ensures that computationally expensive operations, such as data materialization, embedding generation, and index construction are performed asynchronously, while the runtime path remains lightweight and latency-optimized.

\begin{figure}[t]
    \centering
    \includegraphics[width=1\linewidth]{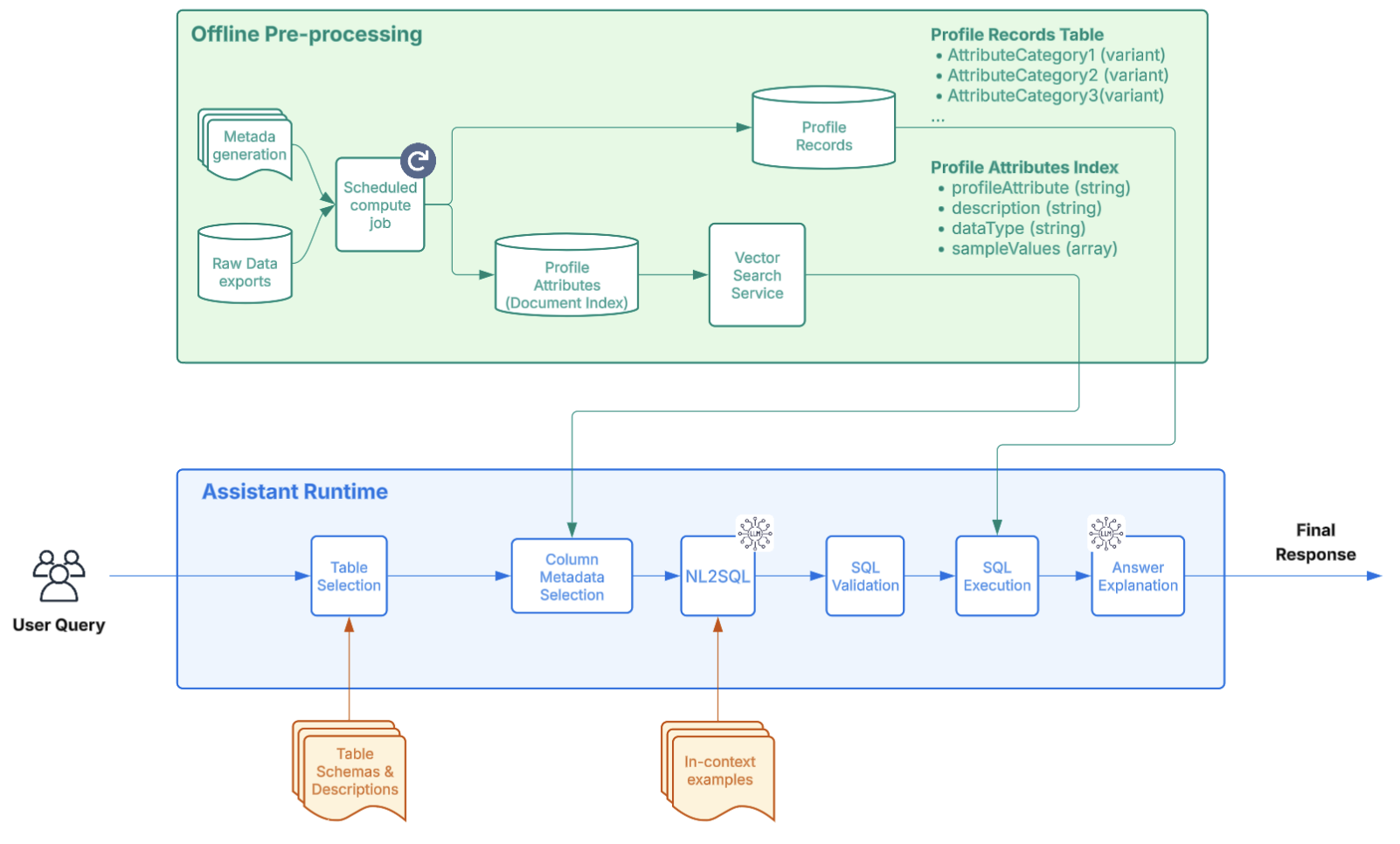}
    \caption{High-level system architecture of REALMS: offline pre-processing and real-time AI Assistant runtime.}
    \label{fig:architecture}
\end{figure}

\subsection{Offline Pre-processing}

\begin{figure}[t]
    \centering
    \includegraphics[width=0.5\textwidth]{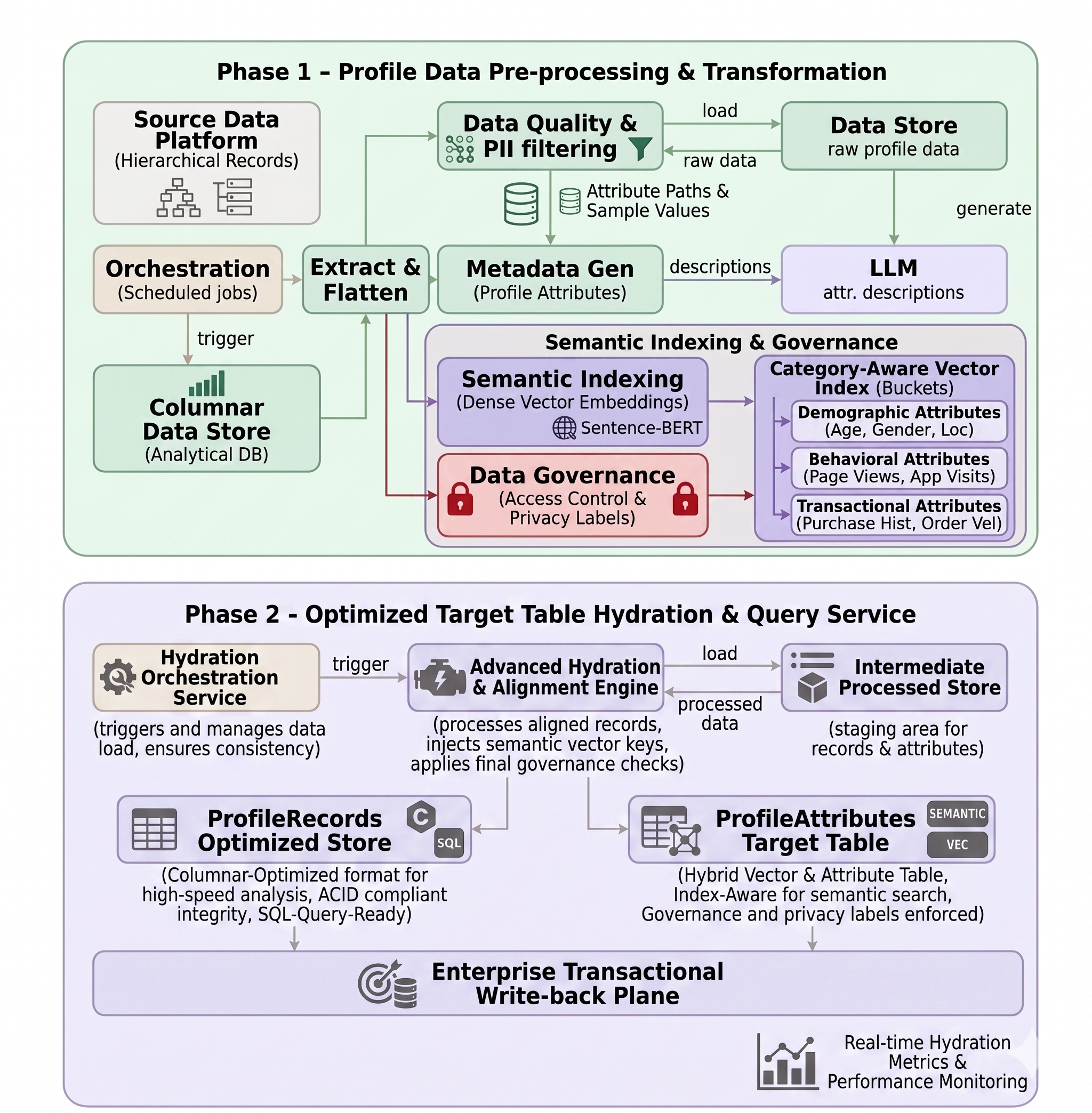}
    \caption{Offline pre-processing and hydration flow.}
    \label{fig:hydration_flow}
\end{figure}

The offline stage transforms raw enterprise profile data into a queryable, indexed representation optimized for real-time retrieval and SQL execution. Figure~\ref{fig:hydration_flow} illustrates the pre-processing pipeline.

Scheduled compute jobs export and flatten hierarchical profile records from the source data platform into a columnar representation stored in a scalable analytical database. This materialization step is necessary because enterprise profiles are typically stored in deeply nested schemas, where direct analytical queries over the raw structures incur prohibitive latency. The columnar representation enables efficient predicate evaluation and aggregation while preserving the full attribute space.

Concurrently, the system generates attribute-level metadata for each schema element, including natural-language descriptions derived from attribute paths, data types, and sample values. Dense vector embeddings are computed for each attribute description using a sentence embedding model~\cite{reimers2019sentence}. Attributes are then organized into \emph{categorical buckets} based on their semantic type, for example, demographic attributes (age, gender, location), behavioral attributes (page views, app visits), and transactional attributes (purchase history, order value). Each bucket is independently indexed in a vector search service to support parallel, category-aware retrieval. This categorical organization addresses a key limitation of na\"ive dense retrieval over large schemas: enterprise schemas often contain hundreds of semantically similar attributes (e.g., multiple date fields or numeric counters), and without categorical separation, retrieval results tend to concentrate within a single semantic cluster, reducing coverage of the attribute space and degrading downstream SQL generation accuracy. Additionally, data governance labels from the platform schema are ingested during pre-processing to annotate attributes with access control and privacy policies, enabling compliance enforcement at query time.

\subsection{AI Assistant Runtime Query Pipeline}

\begin{figure}[t]
    \centering
    \includegraphics[width=1\linewidth]{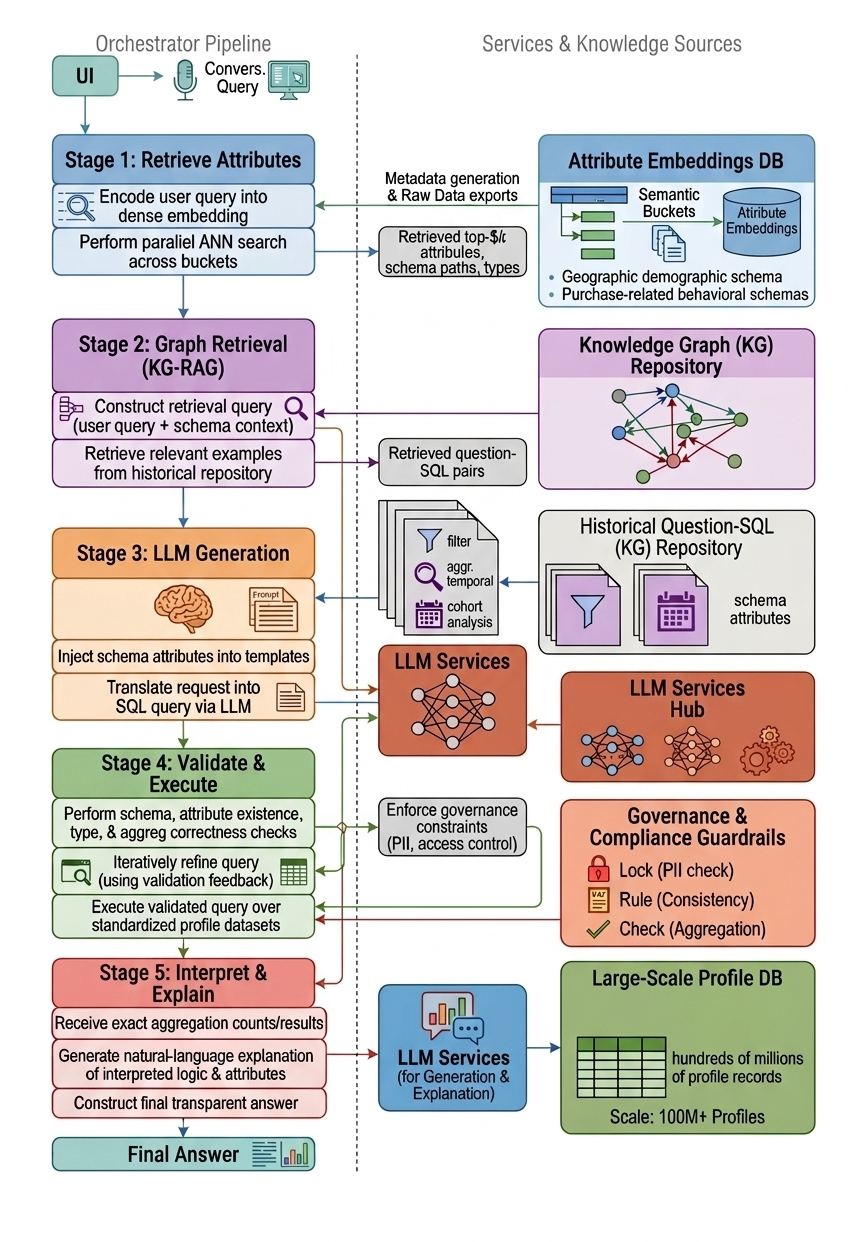}
    \caption{Five stages demonstration of the REALMS runtime query processing pipeline.}
    \label{fig:5_stage}
\end{figure}

\begin{figure}
    \centering
\includegraphics[width=0.9\linewidth]{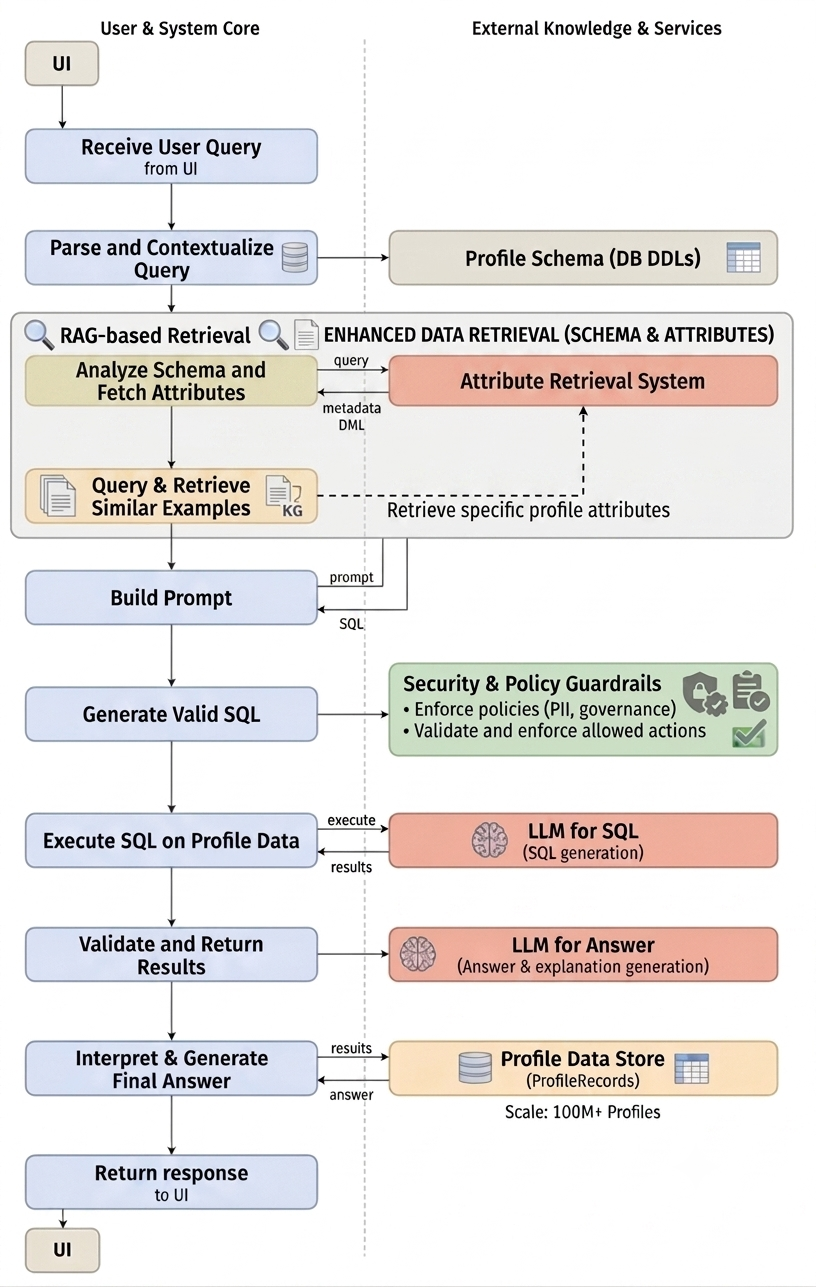}
    \caption{End-to-end REALMS runtime flow.}
    \label{fig:runtime_flow}
\end{figure}

At inference time, REALMS transforms a natural-language audience request into an executable analytical query through a five-stage runtime pipeline and end-to-end runtime flow, illustrated in Figures ~\ref{fig:5_stage} and~\ref{fig:runtime_flow}. The pipeline combines schema-aware retrieval, graph-based retrieval-augmented generation, LLM-powered semantic parsing, scalable query execution, and result interpretation to support accurate and interactive audience sizing over large-scale enterprise profile datasets. 

\textbf{Schema-Aware Attribute Retrieval.} Given a user query, REALMS first identifies the subset of schema attributes most relevant to the requested audience definition. The query is encoded into a dense embedding using the same model employed during offline indexing~\cite{karpukhin2020dense}. The embedding is then used to perform parallel approximate nearest-neighbor searches across all attribute buckets. Unlike global retrieval approaches that may overemphasize a single attribute category, REALMS retrieves the top-$k$ candidate attributes from each semantic bucket, ensuring balanced coverage across heterogeneous profile dimensions.

For example, given the query \emph{``How many users in California made a purchase in the last 30 days?''}, the retrieval stage identifies geographic attributes from demographic schemas and purchase-related attributes from behavioral schemas. The retrieved attributes, together with their schema paths, descriptions, and data types, are assembled into a structured schema context that grounds downstream query generation.

\textbf{Knowledge-Graph Retrieval-Augmented Generation (KG-RAG).} After identifying the relevant schema attributes, REALMS retrieves semantically and structurally similar examples from a repository of historical audience requests and their validated SQL implementations. Rather than maintaining a flat collection of examples, REALMS organizes this repository as a query knowledge graph, where nodes represent natural-language audience definitions, schema attributes, SQL query templates, operators, and aggregation patterns, while edges capture semantic, structural, and execution-level relationships among them.

Given a new request, REALMS performs retrieval over the knowledge graph using both the user query and the retrieved schema context. The retrieval process identifies the top-$k$ question-SQL pairs that are most relevant to the target analytical intent and query structure. This graph-based retrieval strategy differs from conventional RAG systems that rely solely on semantic similarity in a vector index. By explicitly modeling relationships among user intents, schema elements, and SQL patterns, the knowledge graph enables retrieval of examples that are not only semantically related but also structurally aligned with the target query.

The retrieved examples serve as retrieval-augmented context for downstream SQL generation. Because enterprise audience-sizing requests frequently exhibit recurring analytical patterns, such as demographic filtering, behavioral segmentation, temporal constraints, cohort construction, and aggregation operations, grounding generation in previously validated SQL examples significantly improves generation accuracy, reduces hallucinations, and promotes consistent query construction across heterogeneous schemas.

\textbf{LLM-Powered NL2SQL Generation.} Using both the schema context and the retrieved examples, REALMS translates the user request into an executable SQL query. The generation prompt consists of three components: (1) the original natural-language request, (2) the retrieved schema attributes and metadata, and (3) the top-$k$ retrieved question-SQL demonstrations obtained from the KG-RAG stage.

In addition to retrieval-augmented examples, REALMS employs a library of template-based prompt structures that capture common analytical patterns such as filtering, aggregation, ranking, counting, percentage computation, and cohort analysis. These templates remain independent of any organization-specific schema and operate through placeholder-based attribute references. During inference, retrieved schema attributes are injected into the prompt, enabling the same prompt framework to generalize across heterogeneous enterprise datasets without customer-specific fine-tuning~\cite{rajkumar2022evaluating, pourreza2023dinsql, gao2024dailsql}.

Guided by both retrieved examples and schema context, the language model generates SQL over the standardized profile representation, resolving attribute references and constructing the appropriate filtering, grouping, aggregation, and temporal operators required by the user request.

\textbf{SQL Validation and Execution.} Since generated queries directly access enterprise data, REALMS performs a validation stage prior to execution. The validator checks schema consistency, attribute existence, type compatibility, aggregation correctness, and governance constraints. In particular, governance enforcement ensures that restricted or personally identifiable information (PII) attributes cannot be exposed through generated queries or query outputs.

Queries that fail validation are not executed. Instead, structured validation feedback is returned to the language model, enabling iterative query refinement in a manner similar to self-correction frameworks proposed in recent NL2SQL systems~\cite{pourreza2023dinsql}. Once validated, the query is executed on a columnar analytical database engine optimized for large-scale aggregation workloads. This execution layer enables exact audience counting and analytical computation over hundreds of millions of profile records while maintaining interactive response latency.

\textbf{Result Interpretation and Explanation.} Finally, REALMS converts execution results into a user-facing response. In addition to returning the requested audience size or aggregate statistic, the system generates a concise natural-language explanation describing the interpreted query logic, selected attributes, filtering criteria, and aggregation operations.

This explanation provides transparency into the system's reasoning process and allows users to verify that the generated query accurately reflects their intent. Furthermore, the explanation serves as a foundation for iterative refinement, enabling users to modify constraints, add additional conditions, or explore related audience segments through subsequent conversational turns. By combining exact execution with interpretable explanations, REALMS delivers both analytical accuracy and usability for non-technical business users.

\section{Evaluation}

We evaluate REALMS along three dimensions: (1) end-to-end query correctness, (2) attribute retrieval component effectiveness, and (3) system latency and scalability under realistic workloads.

\subsection{Evaluation Dataset}

We construct an evaluation dataset consisting of 600 natural-language audience-sizing requests designed to reflect realistic enterprise analytics workloads. The dataset combines two complementary sources.

First, we collect real customer questions observed in production environments. These queries capture authentic business terminology, natural language variations, and practical information needs encountered by marketing and customer-engagement teams.

Second, we supplement the real queries with LLM-generated questions to increase coverage and systematically explore a broader space of linguistic expressions and logical compositions. The generated queries are designed to preserve realistic business semantics while varying attribute combinations, aggregation structures, and constraint formulations.

To evaluate robustness across common analytical tasks, the dataset covers three representative query intents:

(1) Count queries, which return the exact cardinality of profiles satisfying a predicate (e.g., ``How many profiles have visited the mobile application within the last six months and are older than 30 years?'');

(2) Top-$k$ aggregation queries, which identify the highest-ranked groups according to an aggregate metric (e.g., ``What are the top five cities with the highest average number of e-commerce purchases under the specified constraints?'');

(3) Percentage queries, which compute proportions relative to a reference population (e.g., ``What percentage of customers in Segment A reside in California?'').

To further evaluate compositional reasoning, we vary query complexity by controlling the number of referenced attributes (one, two, or three attributes per query). Increasing the number of attributes simultaneously increases the difficulty of schema grounding, logical composition, and SQL generation.

\subsection{System Latency and Scalability}

To evaluate the operational characteristics of REALMS, we conduct load-testing experiments in a staging environment under varying request rates and concurrency levels. For each configuration, we report the total number of requests, failure count, and latency statistics including median, mean, 95th percentile, and 99th percentile response times.

Table~\ref{tab:load_test} summarizes the results. REALMS maintains stable performance under moderate load levels of 10--30 requests per minute (RPM), exhibiting no observed failures and median response times below nine seconds. As workload intensity increases to 60 RPM with ten concurrent users, the system continues to provide interactive response times, achieving a median latency of 10.0 seconds and a 95th-percentile latency of 14.0 seconds. Under this configuration, only two failures are observed among 105 requests, indicating that the system remains operational while approaching the capacity limits of the current deployment.

These results demonstrate that the retrieval, generation, and execution pipeline can support interactive conversational workloads while maintaining predictable latency characteristics across a range of operating conditions.

\begin{table}[t]
\centering
\caption{Load test results under varying RPM and conc. users.}
\label{tab:load_test}
\small
\setlength{\tabcolsep}{3pt}
\renewcommand{\arraystretch}{1.05}
\begin{tabular}{r r r r r r r r}
\hline
\textbf{RPM} & \textbf{Conc.} & \textbf{\#Req} & \textbf{\#Fail} & \textbf{Med (s)} & \textbf{Avg (s)} & \textbf{P95 (s)} & \textbf{P99 (s)} \\
\hline
10 & 1  & 15  & 0 & 5.9  & 6.7  & 11.0 & 11.0 \\
30 & 5  & 68  & 0 & 8.3  & 8.3  & 12.0 & 13.0 \\
60 & 10 & 105 & 2 & 10.0 & 10.5 & 14.0 & 15.0 \\
\hline
\end{tabular}
\end{table}

\subsection{Retrieval and Query Correctness}

REALMS employs schema-aware attribute retrieval to identify a compact set of candidate attributes for downstream SQL generation. This retrieval stage serves two purposes: (1) reducing prompt size to remain within the context constraints of the language model, and (2) improving generation accuracy by grounding the model on the most relevant schema elements.

We evaluate retrieval quality using two metrics. Recall@$k$ measures the fraction of ground-truth attributes that appear among the top-$k$ retrieved candidates. Exact-match rate@$k$ measures the percentage of queries for which all required attributes are successfully retrieved within the top-$k$ results.

For end-to-end query correctness, we compare generated SQL against reference SQL annotations. Because multiple SQL expressions may be semantically equivalent despite syntactic differences, we prioritize execution match, which executes both queries against the underlying analytical database and compares their outputs. When execution-based comparison is unavailable, we additionally report an LLM-as-a-judge metric that assesses semantic equivalence between generated and reference SQL queries.

\begin{table}[t]
\centering
\caption{Performance on the evaluation dataset (accuracy \%).}
\label{tab:eval_results}
\scriptsize
\setlength{\tabcolsep}{4pt}
\renewcommand{\arraystretch}{1.1}
\begin{tabular}{@{}l l r@{}}
\toprule
\textbf{Task} & \textbf{Metric} & \textbf{Score} \\
\midrule
\textbf{Schema attribute retrieval} & Recall@$k$ & 94\% \\
                                     & Exact-match rate@$k$ & 93\% \\
\midrule
\textbf{Structured query correctness} & Execution match & 90\% \\
                                       & LLM-as-a-judge match & 95\% \\
\bottomrule
\end{tabular}
\end{table}

Table~\ref{tab:eval_results} reports performance when $k=5$. REALMS achieves 94\% Recall@$k$ and 93\% Exact-match@$k$, indicating that the retrieval module consistently identifies the attributes required for query construction. This strong retrieval performance translates into high downstream query accuracy, with 90\% execution match and 95\% semantic equivalence according to the LLM-based evaluator. Together, these results demonstrate that schema-aware retrieval and KG-RAG effectively ground SQL generation over large heterogeneous enterprise schemas.

To better understand the behavior of REALMS, we also conduct a series of additional analyses. First, we perform an ablation study to quantify the contribution of schema-aware retrieval, and KG-RAG. Second, we evaluate robustness under increasing query complexity by varying the number of referenced attributes. Finally, we analyze performance across different analytical intents, including count, top-$k$, and percentage queries. Together, these experiments provide a more comprehensive understanding of the factors driving end-to-end query correctness.

\subsubsection{Ablation Study}

\begin{table}[t]
\centering
\caption{Ablation study of REALMS for structured query correctness.}
\label{tab:ablation}
\scriptsize
\setlength{\tabcolsep}{4pt}
\renewcommand{\arraystretch}{1}
\begin{tabular}{lcc}
\toprule
\textbf{Method} &
\textbf{Execution Match (\%)} &
\textbf{LLM-Judge Match (\%)} \\
\midrule
REALMS (Full) & 90\% & 95\% \\
\hspace{2mm}-- w/o KG-RAG & 68\% & 72\% \\
\hspace{2mm}-- w/o Schema Retrieval & 57\% & 63\% \\
\hspace{2mm}-- Random Example Retrieval & 72\% & 75\% \\
\bottomrule
\end{tabular}
\end{table}

Table~\ref{tab:ablation} evaluates the contribution of individual components in REALMS. Removing either schema-aware retrieval or KG-RAG substantially degrades execution accuracy, indicating that both schema grounding and retrieval-augmented examples are critical for robust NL2SQL generation. Randomly retrieved examples provide limited benefit, demonstrating the importance of knowledge-graph-guided retrieval. The validation-and-refinement loop further improves correctness by correcting schema and syntax errors prior to execution.

\subsubsection{Query Complexity Breakdown}

\begin{table}[t]
\centering
\caption{Performance under different query complexities.}
\label{tab:complexity}
\scriptsize
\setlength{\tabcolsep}{6pt}
\renewcommand{\arraystretch}{1.1}
\begin{tabular}{lccc}
\toprule
\textbf{\# Attributes} &
\textbf{Recall@5 (\%)} &
\textbf{Execution Match (\%)} &
\textbf{LLM-Judge Match (\%)} \\
\midrule
1 Attribute & 100\% & 100\% & 99\% \\
2 Attributes & 98\% & 94\% & 97\% \\
3 Attributes & 93\% & 88\% & 93\% \\
\bottomrule
\end{tabular}
\end{table}

Table~\ref{tab:complexity} reports performance as a function of query complexity. As the number of referenced attributes increases, the difficulty of schema grounding and logical composition grows. Nevertheless, REALMS maintains strong retrieval and execution accuracy, demonstrating robustness to increasingly complex audience definitions.

\subsubsection{Query Intent Breakdown}

\begin{table}[t]
\centering
\caption{Performance across query intents.}
\label{tab:intent}
\scriptsize
\setlength{\tabcolsep}{6pt}
\renewcommand{\arraystretch}{1.1}
\begin{tabular}{lccc}
\toprule
\textbf{Intent} &
\textbf{Recall@5 (\%)} &
\textbf{Execution Match (\%)} &
\textbf{LLM-Judge Match (\%)} \\
\midrule
Count & 96\% & 95\% & 97\% \\
Top-$k$ & 95\% & 88\% & 90\% \\
Percentage & 93\% & 92\% & 93\% \\
\bottomrule
\end{tabular}
\end{table}

Table~\ref{tab:intent} breaks down performance by analytical intent. REALMS achieves consistently strong results across count, top-$k$, and percentage queries, indicating that the proposed retrieval and generation pipeline generalizes beyond simple counting tasks and supports a diverse range of audience analytics workloads.

\section{Conclusion}

We present REALMS, a real-time conversational system for \emph{exact} audience sizing over large-scale enterprise profile stores. REALMS unifies schema-aware attribute retrieval, LLM-driven NL2SQL generation, and a standardized execution layer to handle heterogeneous high-dimensional schemas, nested data, and strict interactive latency constraints. Experiments on real enterprise datasets demonstrate strong retrieval quality, high SQL execution accuracy, and low end-to-end latency, enabling audience insights in seconds. More broadly, REALMS exemplifies conversational analytics in which an LLM acts as a semantic query planner that links information retrieval with database execution for structured aggregation.

\bibliographystyle{plain}
\bibliography{paper_ref}

\end{document}